\documentclass[10pt,twocolumn,letterpaper]{article}

\usepackage{iccv}
\usepackage{times}
\usepackage{epsfig}
\usepackage{graphicx}
\usepackage{amsmath}
\usepackage{amssymb}
\usepackage{subcaption}
\usepackage{multirow}
\usepackage{romannum}
\usepackage{authblk}
\usepackage{blindtext}
\makeatletter
\renewcommand\AB@affilsepx{, \protect\Affilfont}
\makeatother
\graphicspath{{fg/}}
\usepackage{mathtools}
\newcommand\aug{\fboxsep=-\fboxrule\!\!\!\fbox{\strut}\!\!\!}

\usepackage[breaklinks=true,bookmarks=false]{hyperref}

\iccvfinalcopy 

\def\iccvPaperID{2248} 

\ificcvfinal\fi

\begin{document}

\title{Human Attention in Image Captioning: Dataset and Analysis}

\author[1]{Sen He}
\author[2,3]{Hamed R. Tavakoli}
\author[4]{Ali Borji}
\author[1]{Nicolas Pugeault}
\affil[1]{University of Exeter}
\affil[2]{Nokia Technologies}
\affil[3]{Aalto University}
\affil[4]{MarkableAI}


\maketitle
\ificcvfinal\thispagestyle{empty}\fi

\begin{abstract}

In this work, we present a novel dataset consisting of eye movements and verbal descriptions recorded synchronously over images. Using this data, we study the differences in human attention during free-viewing and image captioning tasks. We look into the relationship between human attention and language constructs during perception and sentence articulation. We also analyse attention deployment mechanisms in the top-down soft attention approach that is argued to mimic human attention in captioning tasks, and investigate whether visual saliency can help image captioning. Our study reveals that (1) human attention behaviour differs in free-viewing and image description tasks. Humans tend to fixate on a greater variety of regions under the latter task, (2) there is a strong relationship between described objects and attended objects ($97\%$ of the described objects are being attended), (3) a convolutional neural network as feature encoder accounts for human-attended regions during image captioning to a great extent (around $78\%$), (4) soft-attention mechanism differs from human attention, both spatially and temporally, and there is low correlation between caption scores and attention consistency scores. These indicate a large gap between humans and machines in regards to top-down attention, and (5) by integrating the soft attention model with image saliency, we can significantly improve the model's performance on Flickr30k and MSCOCO benchmarks. The dataset can be found at: \url{https://github.com/SenHe/Human-Attention-in-Image-Captioning}.
\end{abstract}

\section{Introduction}

\begin{figure}
\centering
\includegraphics[width=0.5\textwidth]{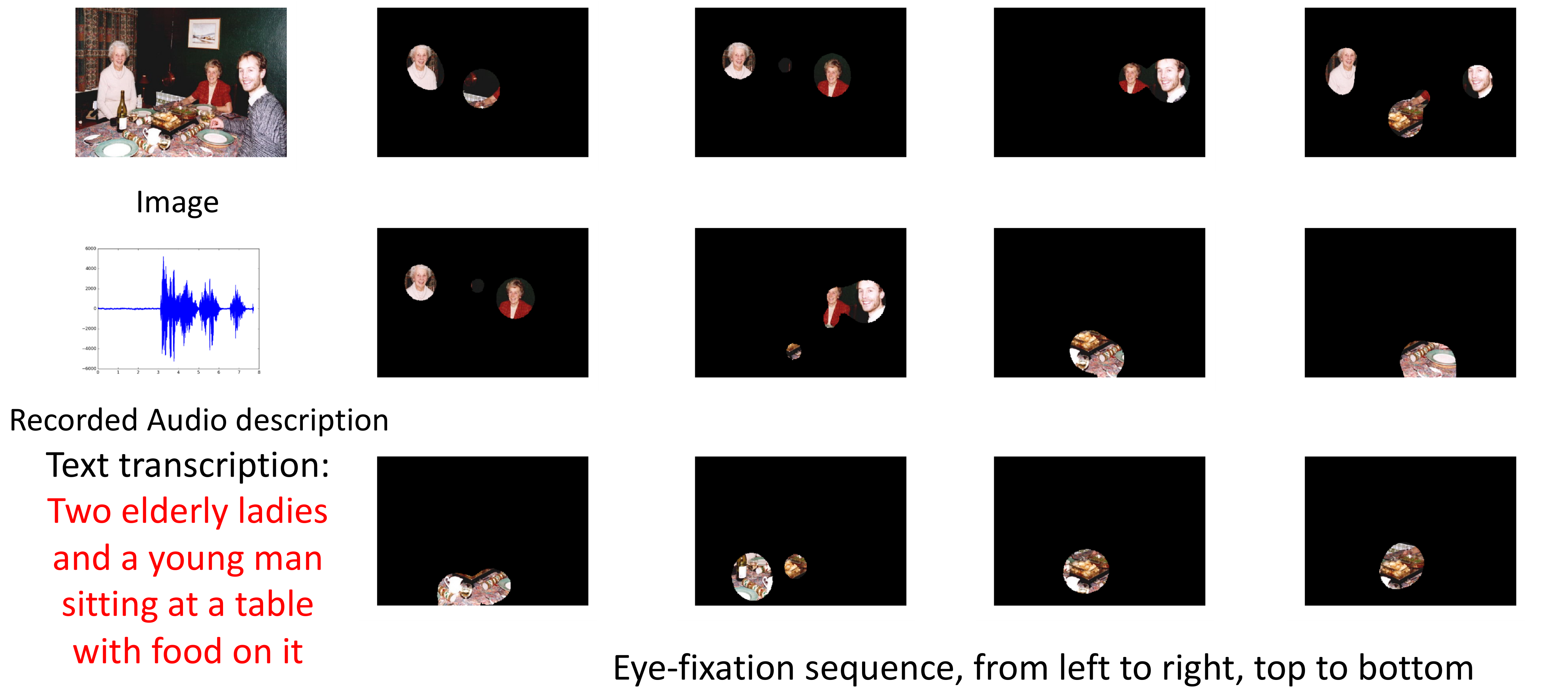}
\vspace{-15pt}
\caption{An example of the data collected in our dataset, including the image shown to the subject, the subject's audio description for this image, the textual transcription of this description, and the sequence of eye-fixations while the subject watched and described the image.}
\vspace{-10pt}
\label{fig:data}
\end{figure}
\begin{quote}
``Two elderly ladies and a young man sitting at a table with food on it.''
\end{quote}
This sentence is an example of how someone would describe the image in Fig.~\ref{fig:data}. Describing images in a few words and extracting the gist of the scene while ignoring unnecessary details is an easy task for humans, that can in some cases be achieved from only a very brief glance.
In a stark contrast, providing a formal algorithm for the same task is an intricate challenge that has been beyond the reach of computer vision for decades. 
Recently, with the availability of powerful deep neural network architectures and large scale datasets, new data-driven approaches have been proposed for automatic captioning of images and have demonstrated intriguing performance~\cite{chen2017sca,lu2017knowing,vinyals2015show,xu2015show}. Although there is no proof that such models can fully capture the complexity of visual scenes, they appear to be able to produce credible captions for a variety of images. 
This raises the question of whether such artificial systems are using similar strategies employed by the human visual system to generate captions. 



One clue onto how humans perform the captioning task is through the study of visual attention via eye-tracking. Attention mechanisms have been studied from different perspectives under the umbrella terms of visual attention (bottom-up and top-down mechanisms of attention), saliency prediction (predicting fixations), as well as eye movement analysis. A large number of studies in computer vision and robotics have tried to replicate these capabilities for different applications such as object detection, image thumbnailing, and human-robot interaction~\cite{borji2018saliency,borji2013state,borji2012quantitative}. There has been a recent trend in adopting attention mechanisms for automatic image captioning~\eg ~\cite{chen2017sca,lu2017knowing,xu2015show}. Such research papers often show appealing visualizations of feature importance over visual regions accompanied with the corresponding phrase ``mimicking human attention". One may ask, ``Is this really the same as human attention?'' and ``How much such mechanisms agree with human attention during describing content?''.


 In this work, we strive to answer the aforementioned questions. We establish a basis by studying how humans attend to scene items under the captioning task. 
 
 Our contributions include: i) introducing a dataset with synchronously recorded eye-fixations and scene descriptions (in verbal form), which provides the largest number of instances at the moment, ii) comparing human attention during scene free-viewing with human attention during describing images, iii) analyzing the relationship between eye-fixations and descriptions during image captioning, iv) comparing human attention and machine attention in image captioning, and v) integrating image saliency with soft attention to boost image captioning performance.
 


\section{Related Work}

\subsection{Bottom-up attention and saliency prediction} Predicting where humans look in an image or a video is a long standing problem in computer vision, a review of which is outside the scope of this manuscript (See~\cite{borji2013state}). We review some of the recent works in bottom-up attention modeling in the following. 
Currently, the most successful saliency prediction models rely on deep neural architectures. Salicon~\cite{jiang2015salicon} is the largest dataset (10k training images) for saliency prediction in free-viewing. Based on the Salicon dataset, the SAM model~\cite{cornia2018sam} uses an LSTM~\cite{hochreiter1997long} network, which can attend to different salient regions in the image. Deep gaze \Romannum{2}~\cite{Kummerer_2017_ICCV} uses features from different layers of a pre-trained deep model and combines them with the prior knowledge (center-bias) to predict saliency. He~\etal~\cite{he2019understanding} analysed the inner representations learned by deep saliency models.
These models are trying to replicate the bottom-up attention mechanism of humans during free-viewing of natural scenes. 

\subsection{Neural image captioning}
The image captioning task can be seen as a machine translation problem, \eg translating an image to an English sentence. A breakthrough in this task has been achieved with the help of large scale databases for image captioning (\eg Flickr30k~\cite{young2014image}, MSCOCO~\cite{lin2014microsoft}) that contain a large number of images and captions (\ie, source and target instances). The neural captioning models often consist of a deep Convolutional Neural Network (CNN) and a Long Short Term Memory (LSTM)~\cite{hochreiter1997long} language model, where the CNN part generates the feature representation for the image, and the LSTM cell acts as a language model, which decodes the features from the CNN part to the text, \eg~\cite{vinyals2015show}. In this paper, we mainly focus on models that incorporate attention mechanisms. Xu~\etal~\cite{xu2015show} introduce a soft-attention mechanism to the approach in~\cite{vinyals2015show}. That is, during the generation of a new word, based on the previously generated word and the hidden state of the language model, their model learns to put spatial weight on the visual features. Instead of re-weighting features only spatially, Chen \etal~\cite{chen2017sca} exploit spatial and channel wise weighting. Lu~\etal~\cite{lu2017knowing} utilize memory to prevent the model in attending mainly to the visual content and enforce it to utilize textual context as well. This is referred to as \textit{adaptive attention}. Chen~\etal~\cite{chen2018boosted} apply the visual saliency to boost the captioning model. They use weights learned from saliency prediction to initialize their captioning model, but the relative improvement is marginal compared to training their model from scratch.

\subsection{Human attention and image descriptions}
In the vision community, some previous works have investigated the relationship between human attention and image captioning (\eg~\cite{tanenhaus1995integration,Itti_Arbib06arl}). Yun~\etal~\cite{yun2013studying} studied the relationship between gaze and descriptions, where the human gaze was recorded under the free-viewing condition. In their work, subjects were shown an image for 3 seconds, and another group of participants described the image content separately. We will refer to their data as \textit{sbugaze}. Tavakoli~\etal~\cite{tavakoliy2017paying} pushed further to investigate the relation between machine-generated and human-generated descriptions. They looked into the contribution of boosting visual features spatially using saliency models as a replicate to bottom-up attention. Abhishek~\etal~\cite{das2017human} studied the relationship between human attention and machine attention in visual question answering. Contrary to previous studies, we focus on the human attention under the image captioning task and investigate the attention behaviour by human and machine.

In the natural language community, eye-tracking and image descriptions have been used to study the cause of ambiguity between languages, \eg English vs Dutch~\cite{miltenburg2018didec}. Vaidyanathan \etal~\cite{vaidyanathan2018snag} investigated the relation between linguistic labels and important regions in the image by utilizing eye tracking data and image descriptions. 
In contrast to existing datasets in the natural language processing community, our dataset features a higher number of instances and images in total, making it more suitable for vision related tasks (see Table~\ref{tab:data_com} for a comparison). In contrast to prior works, we also pursue a different goal which is: \textit{understanding how well current computational attention mechanisms in image captioning models align with human attention behaviour during image description task}.

\begin{table}[t]
\small
\caption{Comparing our data with other similar datasets.}
\centering
\small
\begin{tabular}{l | c c c}
\hline
Dataset      &  \# images   & \# subjects  &\# Instances \\
\hline
\hline
DIDEC~\cite{miltenburg2018didec}     & 305       &  45    &   4604 \\
SNAG~\cite{vaidyanathan2018snag}    &   100        & 30      & 3000 \\
sbugaze~\cite{yun2013studying} & 1000 & 3 & 3000 \\
Ours   &   4000     & 16         & 14000\\
\hline
\end{tabular}
\label{tab:data_com}
\vspace{-10pt}
\end{table}

\section{Data Collection}

\paragraph{Stimuli:} Our collected data is organised in two corpora, denoted by \textit{capgaze1} and \textit{capgaze2} respectively. The \textit{capgaze1} corpus is used for analysis in the paper and the \textit{capgaze2} corpus is used for modelling the visual saliency under the image captioning task. 
For \textit{capgaze1}, 1,000 images were selected from the Pascal-50S dataset~\cite{vedantam2014collecting}, which provides 50 captions per image by humans and annotated semantic masks with 222 semantic categories (the same images as in \textit{sbugaze}). 
For \textit{capgaze2}, 3,000 images were randomly chosen from the MSCOCO~\cite{lin2014microsoft}. Yun~\etal~\cite{yun2013studying} recorded eye movements of subjects during free-viewing images in Pascal-50S. Thus, we can use \textit{capgaze1} to compare human attention under free-viewing or captioning.

\paragraph{Apparatus:} Precise recording of subjects' fixations in the image captioning task requires specialized accurate eye-tracking equipment, making crowd-sourcing impractical for this purpose. We used a \textit{Tobii X2-30} eye-tracker to record eye movements under the image captioning task in a controlled laboratory condition. The eye-tracker was positioned at the bottom of the laptop screen with a resolution of $1920 \times 1080$. The subject's distance from the screen was about $40$cm. Subject was asked to simultaneously look at the image and describe it in one sentence in verbal form. The eye-tracker and an embedded voice recorder in the computer recorded the subject's eye movements and descriptions synchronously for each image. 

Five subjects (postgraduate students, native English speakers, 3 males and 2 females) participated in the data collection for \textit{capgaze1} corpus. All five subjects finished the data collection over all 1,000 images in this corpus. Eleven subjects (postgraduate students, 3 females and 8 males) participated in the data collection over \textit{capgaze2} corpus. Each image in this corpus has the recorded data from three different subjects. The image presentation order was randomized across subjects. For each subject, we divided the data collection into 20 images per session. Before each session, the eye-tracker was re-calibrated. 
At the start of a session, the subject was asked to fixate on a central red cross, which appeared for $2$s. The image was then displayed on the screen and the subject viewed and described the image. After describing the image, the subject pressed a designated button to move to the next image in the session. An example of the collected data is illustrated in Fig.~\ref{fig:data}. During the experiments, subjects often looked at the image silently for a short while to scan the scene, and then started describing the content spontaneously for several seconds. 

\begin{table}
\setlength{\tabcolsep}{1pt} 
\renewcommand{\arraystretch}{1} 
\small
\caption{Assessing quality of the collected captions against 50 ground-truth captions of the Pascal-50S.} 
\centering
\begin{tabular}{l| c| c}
\hline 
Dataset  & CIDEr  & METEOR \\
  & \ \ mean/variance \ \   & \ \ mean/variance \\
\hline
\hline
sbugaze~\cite{yun2013studying}   &  0.938/0.038                    &          0.368/  0.012 \\
Ours       &  0.937/0.060                    &          0.366/ 0.015\\
\hline
\end{tabular}
\label{tab:data_eva}
\end{table}

\vspace{-10pt}
\paragraph{Post-processing:} After data collection, we manually transcribed the oral descriptions in \textit{capgaze1} corpus into text for all images and subjects. The transcriptions were double-checked and cross checked with the images. We used off-the-shelf part of speech (POS) tagging software~\cite{manning2014stanford} to extract the nouns in the transcribed sentences. We then formed a mapping from the extracted nouns to the semantic categories present in the image. For example, \textit{boys} and \textit{girls} are both mapped into the \textit{person} category.

To check the quality of captions in our collected data, we compute the CIDEr~\cite{vedantam2015cider} and METEOR~\cite{denkowski2014meteor} scores of the collected captions based on the ground truth in \textit{Pascal-50S} dataset (50 sentences for each image). To ensure that eye tracking and simultaneous voice recording have not affected the quality of captions adversely, we compared our scores with the scores of~\textit{sbugaze}~\cite{yun2013studying} captions, that were collected in text form, with asynchronous eye tracking and description collection. Table~\ref{tab:data_eva} summarizes the results, showing that eye-tracking does not appear to distract the subjects as their descriptions rated comparably to the ones in \textit{sbugaze}.

\section{Analysis}
In this section, we provide a detailed analysis of (i) attention during free-viewing and captioning tasks, (ii) the relationship between fixations and generated captions, and (iii) attentional mechanisms in captioning models. 

\subsection{Attention in free-viewing vs. attention in image captioning}

\begin{figure}
\centering
\begin{subfigure}[b]{0.3\linewidth}
\includegraphics[width=\linewidth]{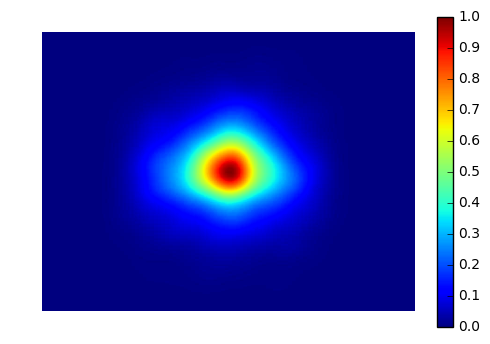}
\caption{\textit{free}}
\end{subfigure}
\begin{subfigure}[b]{0.3\linewidth}
\includegraphics[width=\linewidth]{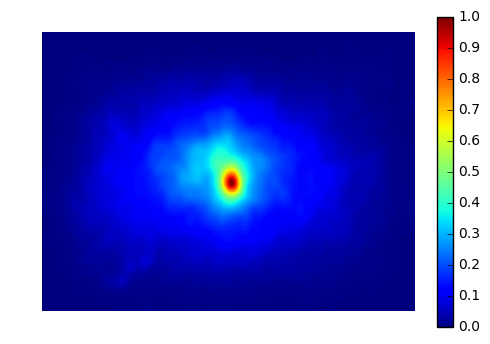}
\caption{\textit{cap3s}}
\end{subfigure}
\begin{subfigure}[b]{0.3\linewidth}
\includegraphics[width=\linewidth]{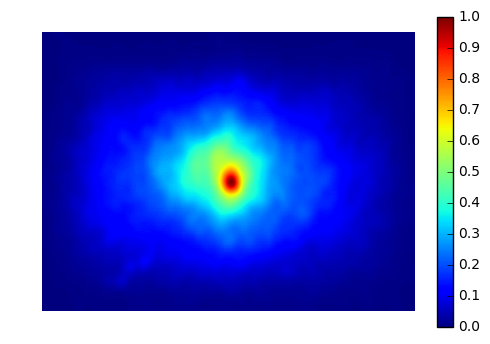}
\caption{\textit{cap}}
\end{subfigure}
\vspace{-5pt}
\caption{Average fixation map across the whole dataset for (a) the free-viewing condition, (b) first 3 seconds of image captioning condition (\textit{cap3s}), and (c) the whole duration of captioning condition (\textit{cap}).}
\label{fig:mean_sal}
\end{figure}

\begin{figure}
\centering
\includegraphics[width=0.4\textwidth]{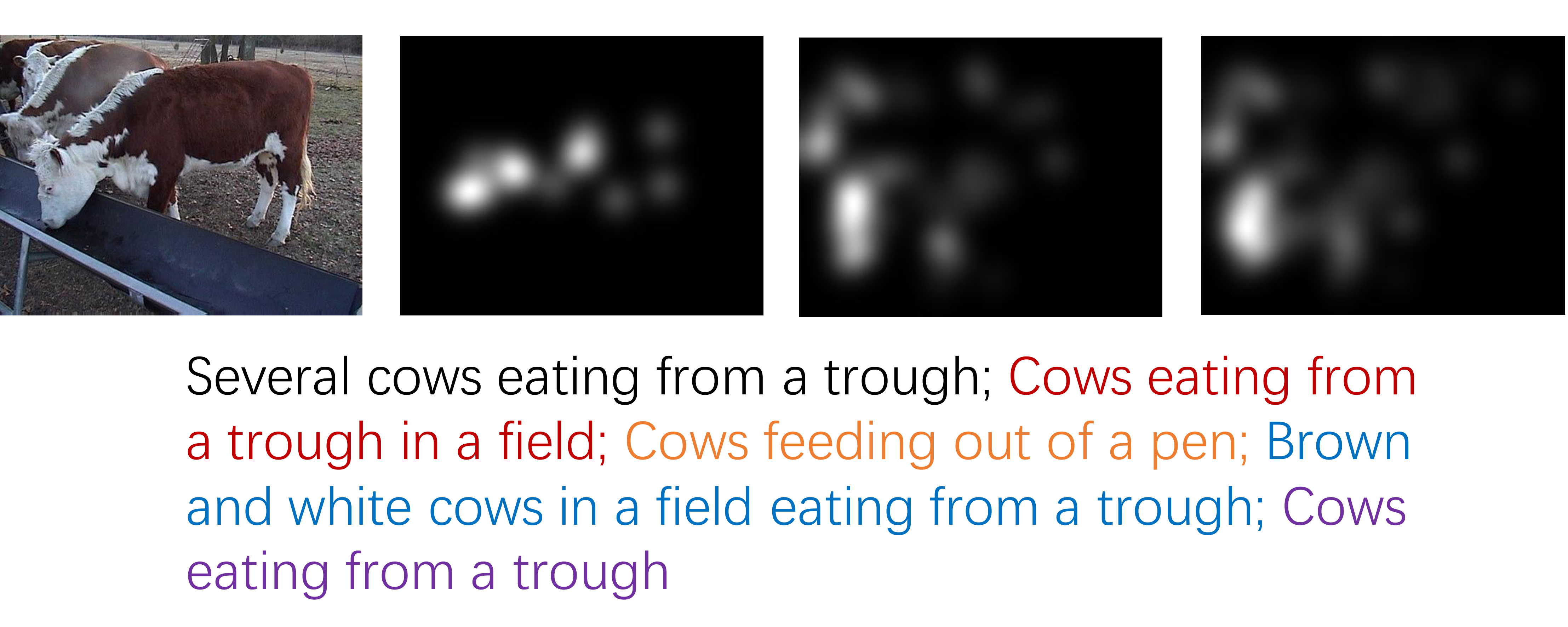}
\vspace{-10pt}
\caption{An example of the difference between fixations in free-viewing and image captioning tasks. From left to right: original image, free-viewing fixations, first 3s fixations, and all fixations in the captioning task. The captions generated by 5 subjects are shown at the bottom.}
\label{fig:at_dif}
\end{figure}


\begin{table}[h!]
\caption{Cross task IOC in terms of AUC-Judd}
\centering
\footnotesize
\begin{tabular}{|c|c |c|c|}
\hline
     &\multicolumn{3}{c |}{\textbf{Reference task}}\\
\cline{2-4}
 &free &cap3s&cap\\
\hline
 free&0.81&0.78&0.75\\
 cap3s&0.84 &0.84&0.81\\
 cap&0.84 &0.85&0.83\\
\hline
\end{tabular}
\label{tab:ioc}
\end{table}

How does attention differ between free-viewing versus describing images?
We first analyze the differences between the two tasks by visualizing the amount of attentional center-bias and the degree of cross task inter-observer congruency (IOC).
The \textit{sbugaze} dataset contains gaze for a maximum of duration of 3s, in free-viewing condition, whereas in our experiments, subjects needed on average $6.79s$ to look and describe each image. To ensure the difference in gaze locations is not solely due to viewing duration, we divide the visual attention in the image captioning task into two cases: i) fixations during the first $3s$ (\textit{cap3s}), and ii) fixations during the full viewing period (\textit{cap}).

The difference in visual attention between free-viewing and image captioning is shown in Figs.~\ref{fig:mean_sal} and \ref{fig:at_dif}. We find that visual attention in free-viewing is more focused towards the central part of the image (\ie high centre-bias), while attention under the image captioning task has higher dispersion over the whole duration of the task.

Table~\ref{tab:ioc} reports the cross task \textit{Inter Observer Congruency} (IOC). To compute cross task IOC, we leave fixations of one subject in one task out and compute its congruency with the fixations of other subjects in another task using the AUC-Judd~\cite{bylinskii2018different} evaluation score. The results shows that human attention in captioning task is different than free-viewing.




\subsection{Analyzing the relationship between fixations and scene descriptions}

How does task-based attention relates to image description?
To answer, we analyze the distribution of fixations on objects in the scene and the relation between attention allocation and noun descriptions in the sentences.
Given described objects ($\mathcal{D}(O)$), non-described objects ($\neg \mathcal{D}(O)$), described background (\eg, mountain, sky, wall) 
denoted as $\mathcal{D}(B)$, non-described background ($\neg \mathcal{D}(B)$), and fixated objects ($F(O)$), we compare the distribution of the fixation data on objects and image background in free-viewing, first $3s$ captioning and full captioning tasks. We compute the attention ratio for regions of interest as:
\begin{equation}
\mbox{attention\_ratio} = \frac{\# \ \text{fixations on a region}}{\# \ \text{total fixations on the image}}
\end{equation}

\begin{figure*}[t]
\centering
\begin{subfigure}[b]{0.32\linewidth}
\includegraphics[width=\linewidth]{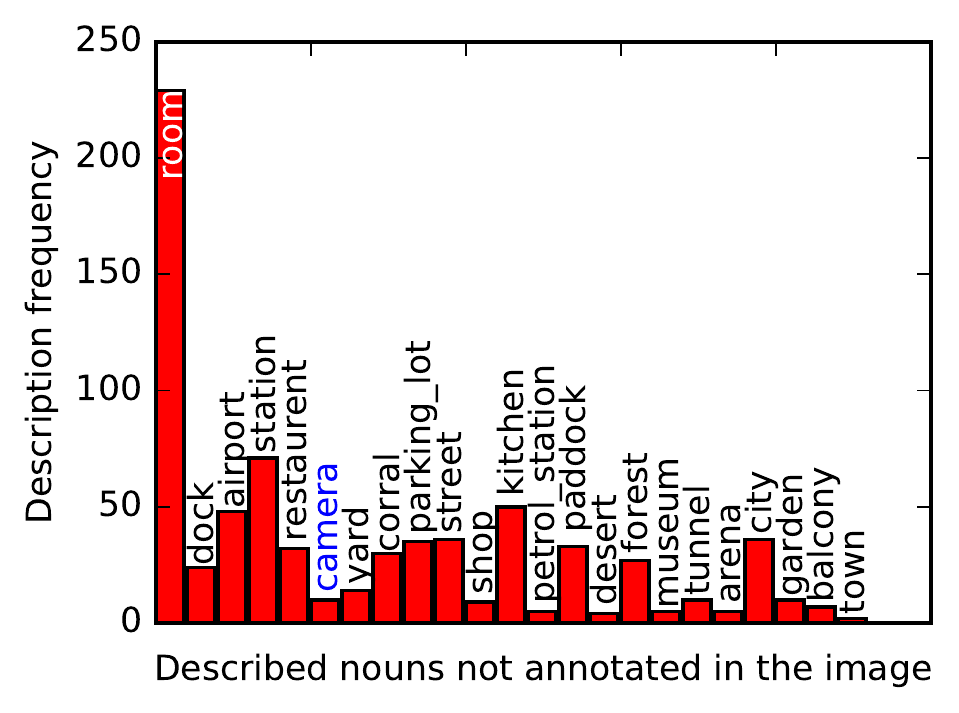}
\end{subfigure}
\begin{subfigure}[b]{0.32\linewidth}
\includegraphics[width=\linewidth]{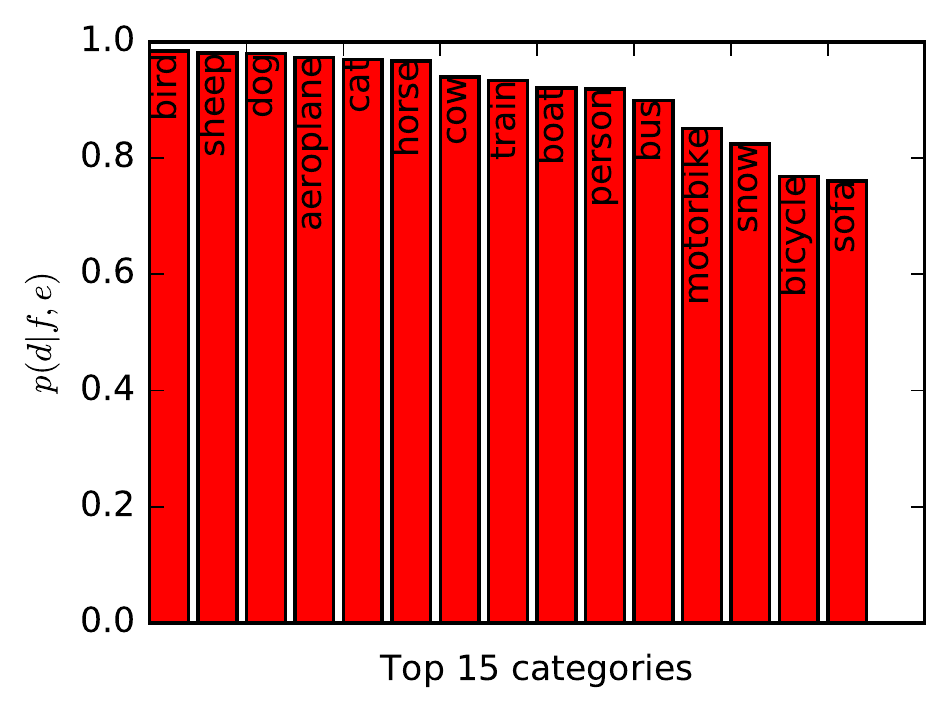}
\end{subfigure}
\begin{subfigure}[b]{0.32\linewidth}
\includegraphics[width=\linewidth]{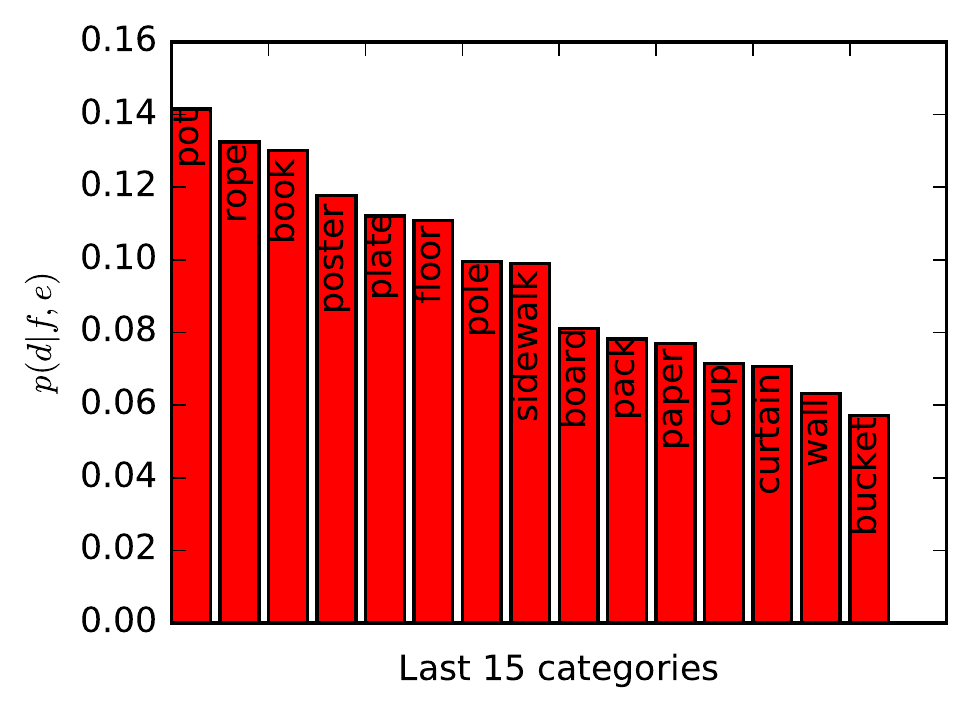}
\end{subfigure}
\vspace{-10pt}
\caption{From left to right: the nouns described in the caption but not annotated in the image; the fixated objects (top 15) that have a very high likelihood to be described; the fixated objects that have a very low likelihood to be described.}
\label{fig:dfn}
\vspace{-10pt}
\end{figure*}

\begin{table}[t]
\small
\caption{Mean attention allocation on different regions (object vs background).}
\centering
\setlength{\tabcolsep}{.5pt} 

\begin{tabular}{c|c|c|c|c }
\hline 
 &$\mathcal{D}(O)$& $\neg \mathcal{D}(O)$&$\mathcal{D}(B)$ &$\neg \mathcal{D}(B)$\\
\hline
\hline
 free&    0.66  & 0.09 & 0.14  & 0.11\\
 \hline
 cap(3s)&    0.68  & 0.09 & 0.14 & 0.09\\
 \hline
 cap&    0.63  & 0.10 & 0.16 & 0.11\\
\hline
\end{tabular}
\label{tab:fixdis}
\end{table}

Are described objects more likely to be fixated? Table~\ref{tab:fixdis} shows the results in terms of overall attention allocation. As depicted, in all viewing conditions most of the fixations correspond to objects that are described in the caption. This is in line with previous findings in~\cite{tavakoliy2017paying}: described objects receive more fixations than background (either described or not) and non-described objects.
When comparing the fixations in the free-viewing and captioning conditions, we see that in the first $3s$ of captioning (the common viewing duration for free-viewing), slightly more attention is allocated to the described objects. Analyzing the captioning task for the full duration, we observe a \textit{decrease} in the attention allocation on described objects and an \textit{increase} in the attention to the described background. This indicates that subjects are more likely to attend to the items which are going to be described in the first few seconds, before shifting their attention towards context-defining elements in the scene.

\begin{table}[t]
\small
\caption{Attention allocation on described objects based on the order in which they appear in the description} 
\vspace{-5pt}
\centering
\begin{tabular}{c | c| c| c| c| c}
&\multicolumn{5}{c }{{Noun Order}}\\
\cline{2-6}
& 1   & 2  & 3  & 4  & 5 \\
\hline
\hline
cap   &  0.486   &0.201   &0.147 &0.097&0.053 \\
\hline
free  &  0.502   &0.204   &0.158 &0.107 & - \\
\hline
\end{tabular}
\label{tab:noun_order}
\end{table}

Are the objects that appear at the start, rather than the end of the description, more likely to be fixated? Table~\ref{tab:noun_order} shows the magnitude of attention allocation to objects with respect to their order of appearance in the descriptions (noun order). We see that nouns that are described first receive a larger fraction of fixations than the subsequent nouns. The slightly lower number in the captioning condition is associated with the change in viewing strategy observed after the first $3s$, as discussed previously.

\begin{table}[t]
\small
\caption{The mean fixation duration ($T$) on described objects vs. non-described objects} 
\centering
\vspace{-5pt}
\begin{tabular}{c | c | c}
 & $\neg \mathcal{D}(O)$  \ \  & $\mathcal{D}(O)$ \\
\hline
\hline
$T_F(O)$ & 0.52 s &1.68s \\
\end{tabular}
\label{tab:dur}
\end{table}

How much time do subjects spend viewing described objects? Synchronous eye tracking and description articulation enables us to investigate the duration of fixations $T_F$ on scene elements, specifically on described and non-described objects. As shown in Table~\ref{tab:dur}, described objects attract longer fixations than non-described objects. This indicates that once an important object grabs the attention, more time is allocated to scrutinize it.

\begin{table}[t] 
\small
\caption{The probability of an object being described when fixated vs. fixated when described.}
\centering
\begin{tabular}{c | c c}
              & $p(\mathcal{D}(O)|F(O),O)$   & $p(F(O)|\mathcal{D}(O),O)$ \\
\hline
\hline
free  &  0.56   & 0.87\\
\hline
cap (3s)  & 0.48  & 0.95 \\
\hline
cap    &  0.44  & 0.96 \\
\hline
\end{tabular}
\label{tab:pfe}
\end{table}

How likely is an object to be described if it is fixated? We compute the probability, $p(\mathcal{D}(O)|F(O),O)$), and compare it with the probability that an object is fixated when it is described (when it is present in the image), $p(F(O)|\mathcal{D}(O),O)$. In other words, are we more likely to fixate on what we describe, or to describe what we fixate?
Results are summarized in Table~\ref{tab:pfe}. They confirm the expectation that described objects are very likely to be fixated, whereas many fixated objects are not described. Interestingly, under the image captioning task, more fixated objects are not described, whereas described objects are more likely to be fixated.


How often do subjects describe something not annotated in the image (\ie, not present in the image at all)? Also, which nouns are described more often and which ones are less likely to be mentioned?
The data is visualized in Fig.~\ref{fig:dfn}. Most occurrences of described but un-annotated nouns are \textit{scene categories} and \textit{places} nouns, that are not annotated as scene elements (because annotations are local and pixel-based). 
One glaring exception to this, where an object not present in the scene is described, is the special case of `camera'. The reference to `camera' is often associated with captions that refer to the \textit{photographer} taking the picture. Since the word camera in this case denotes a property of the scene rather than the material object (here actual camera), we can loosely construe such cases as a scene category.

 


\subsection{Comparing human and machine attention}
How similar are human and machine attention in image captioning? This section describes two analyses performed to answer this question.

\subsubsection{Attention in the visual encoder}
An overlooked aspect in previous research is the amount of saliency that may have been encoded implicitly within the visual encoder of a deep neural network. Consider the situation where a standard convolutional neural network (CNN) architecture, often used for encoding visual features, is used to provide the features to a language model for captioning. We ask (1) to what extend does this CNN capture salient regions of the visual input? and (2) how well do the salient regions of the CNN correspond to human attended locations in the captioning task?

\begin{figure}[t]
\centering
\begin{subfigure}[b]{0.19\linewidth}
\includegraphics[width=\linewidth]{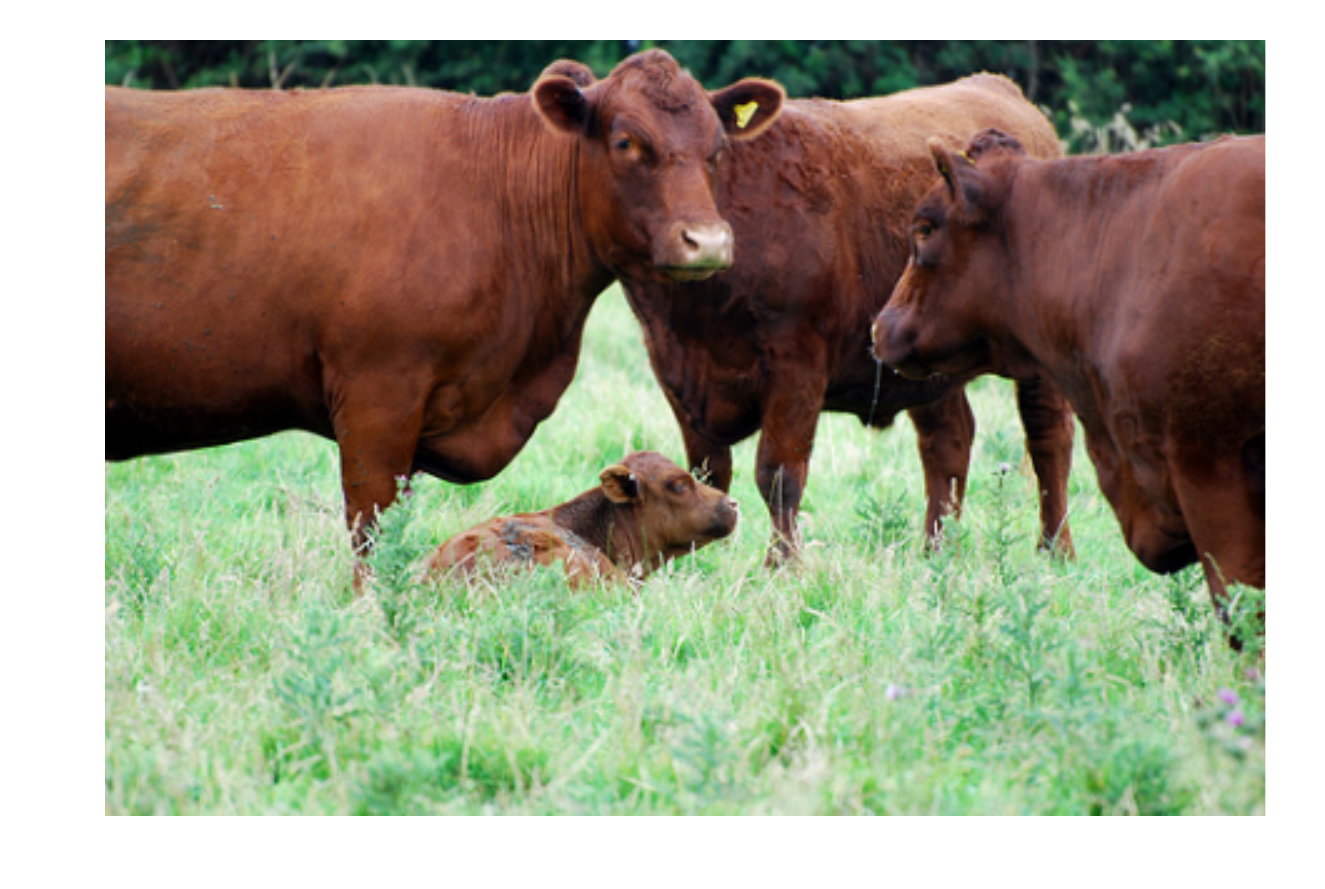}
\end{subfigure}
\begin{subfigure}[b]{0.19\linewidth}
\includegraphics[width=\linewidth]{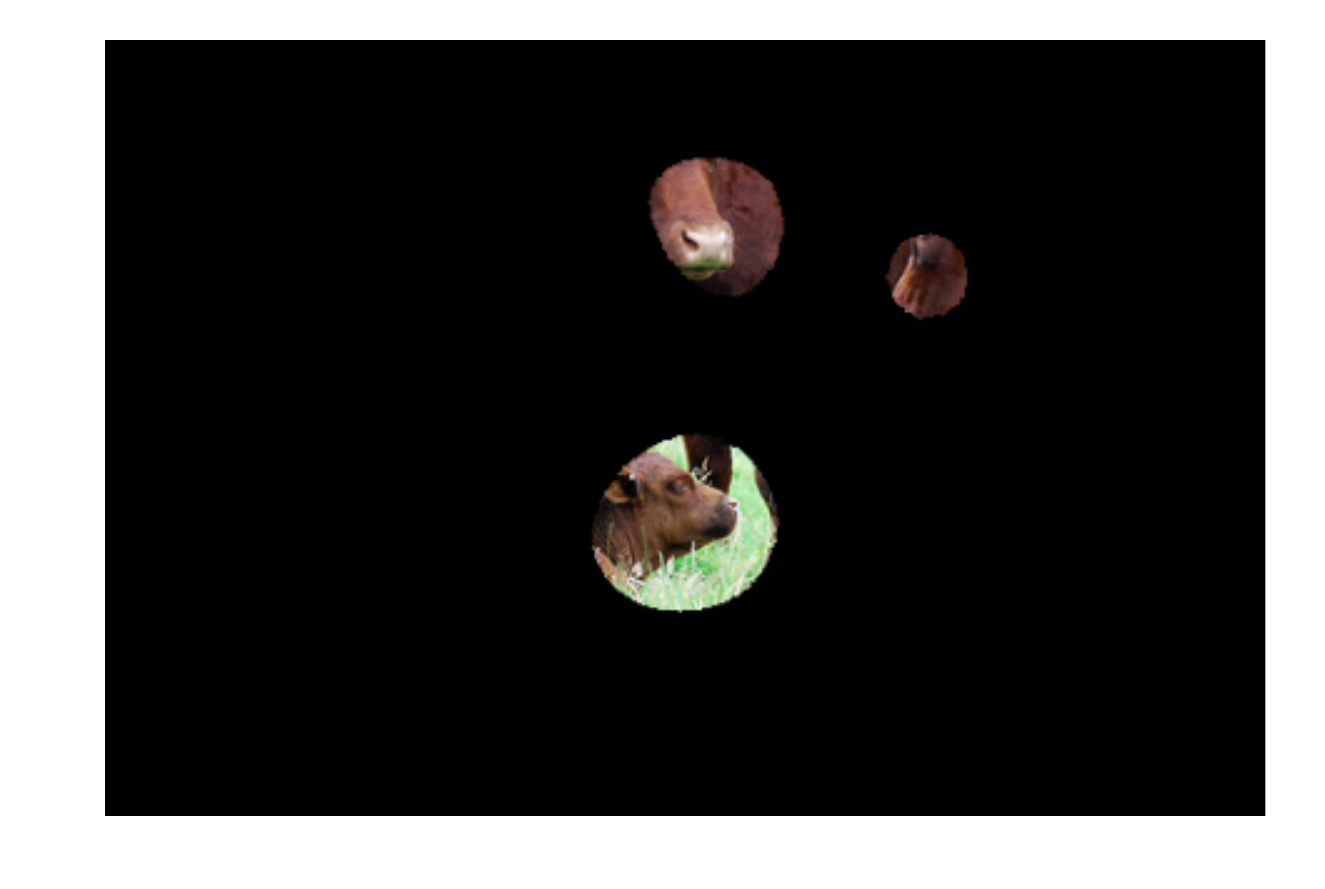}
\end{subfigure}
\begin{subfigure}[b]{0.19\linewidth}
\includegraphics[width=\linewidth]{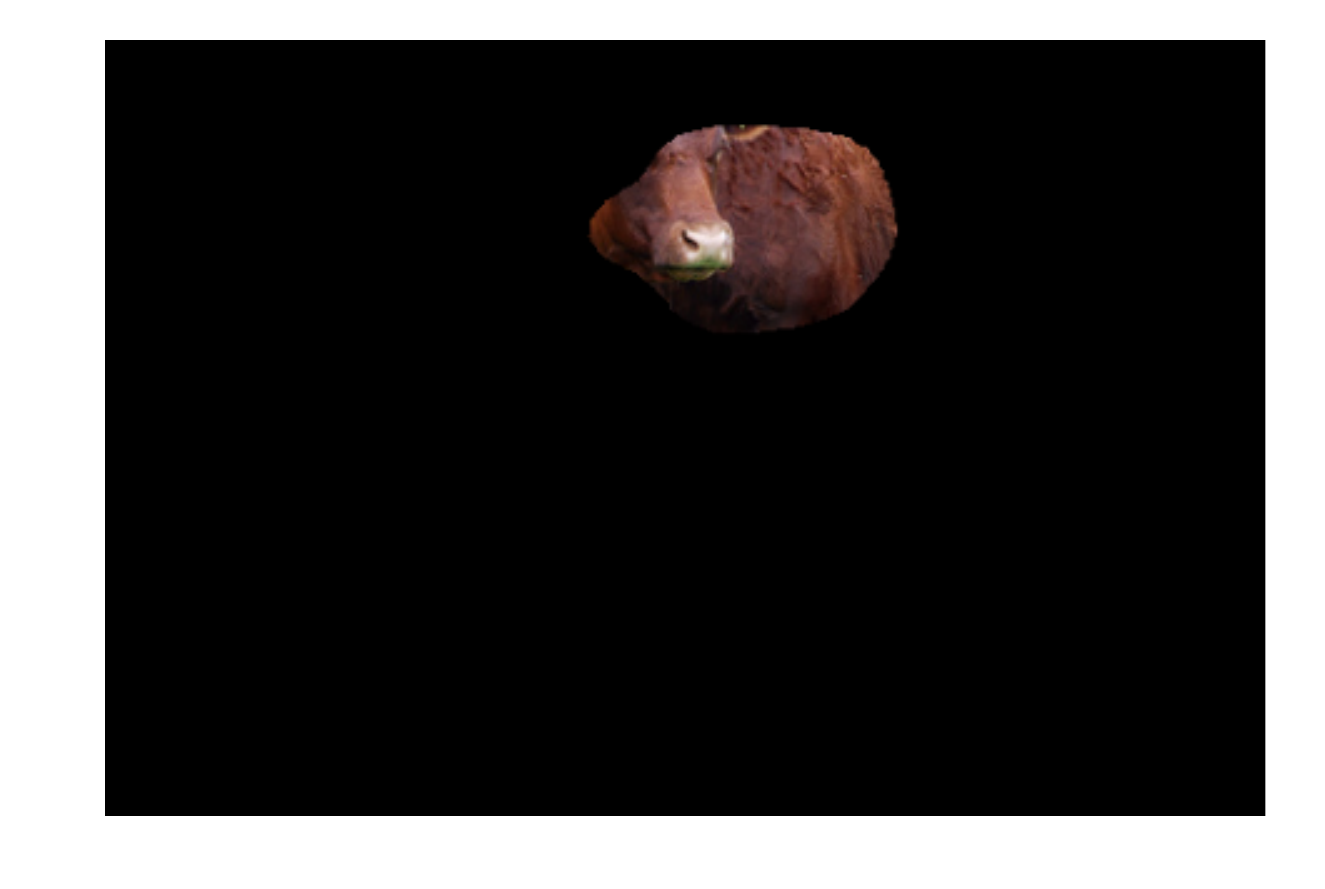}
\end{subfigure}
\begin{subfigure}[b]{0.19\linewidth}
\includegraphics[width=\linewidth]{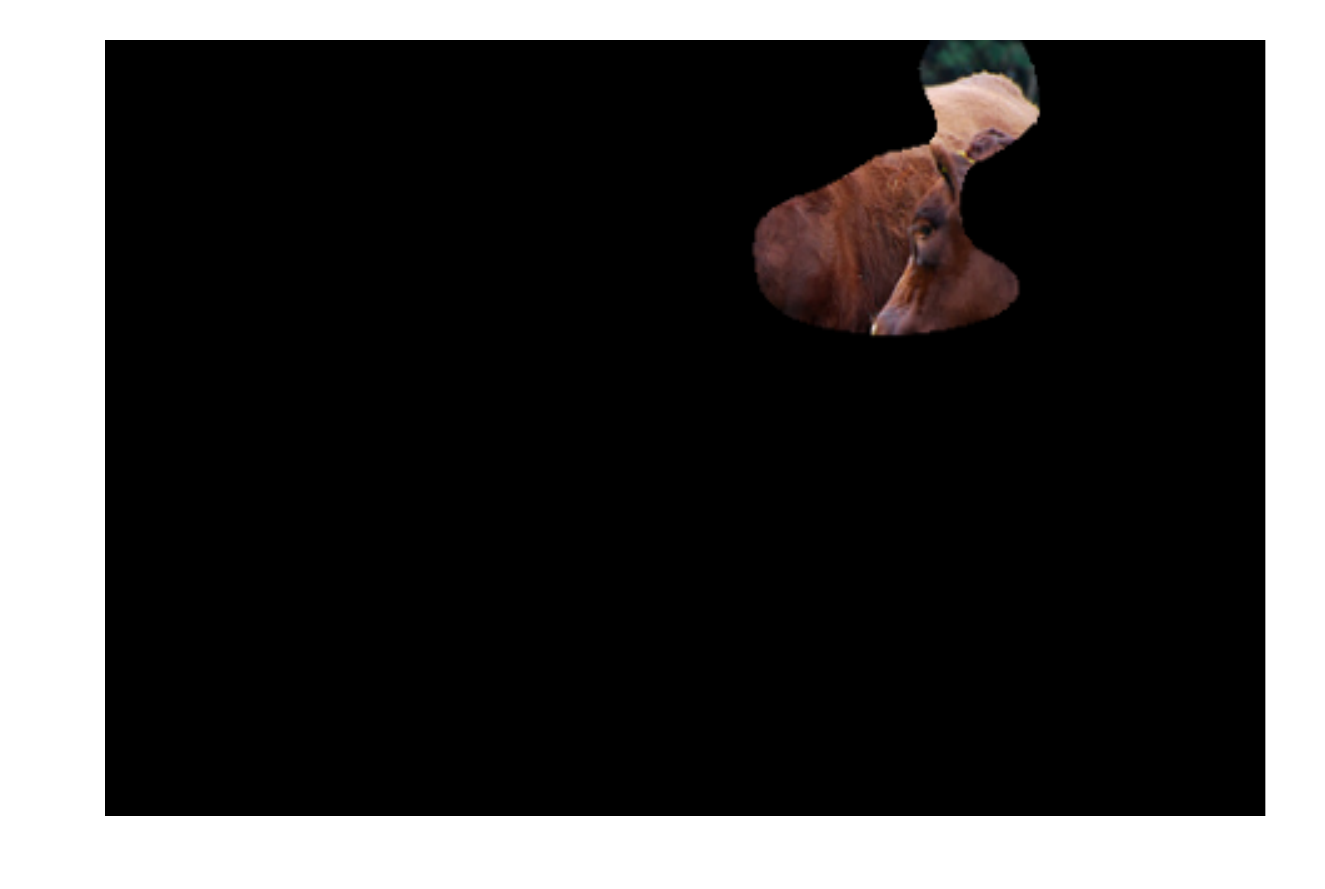}
\end{subfigure}
\begin{subfigure}[b]{0.19\linewidth}
\includegraphics[width=\linewidth]{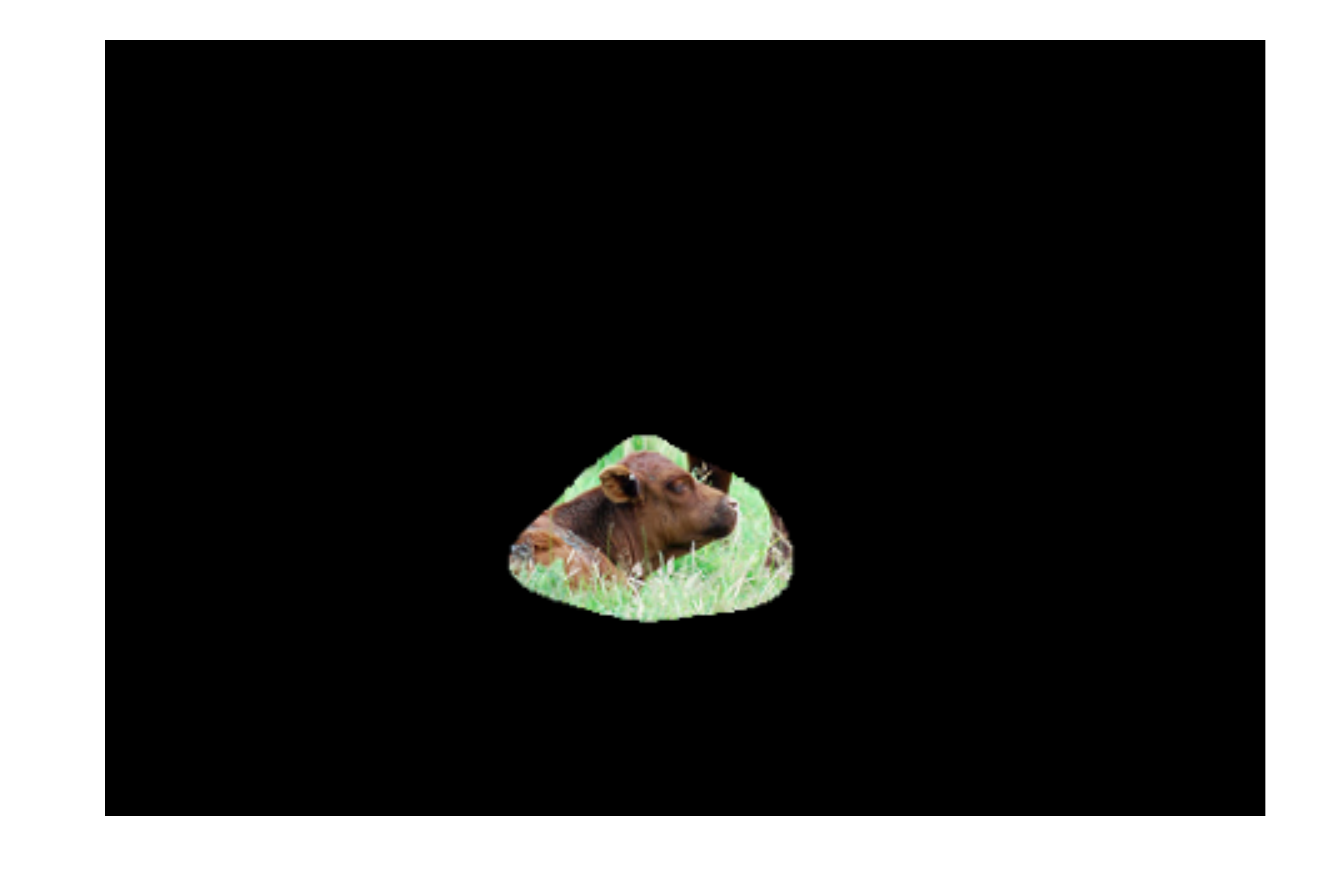}
\end{subfigure}
\begin{subfigure}[b]{0.19\linewidth}
\includegraphics[width=\linewidth]{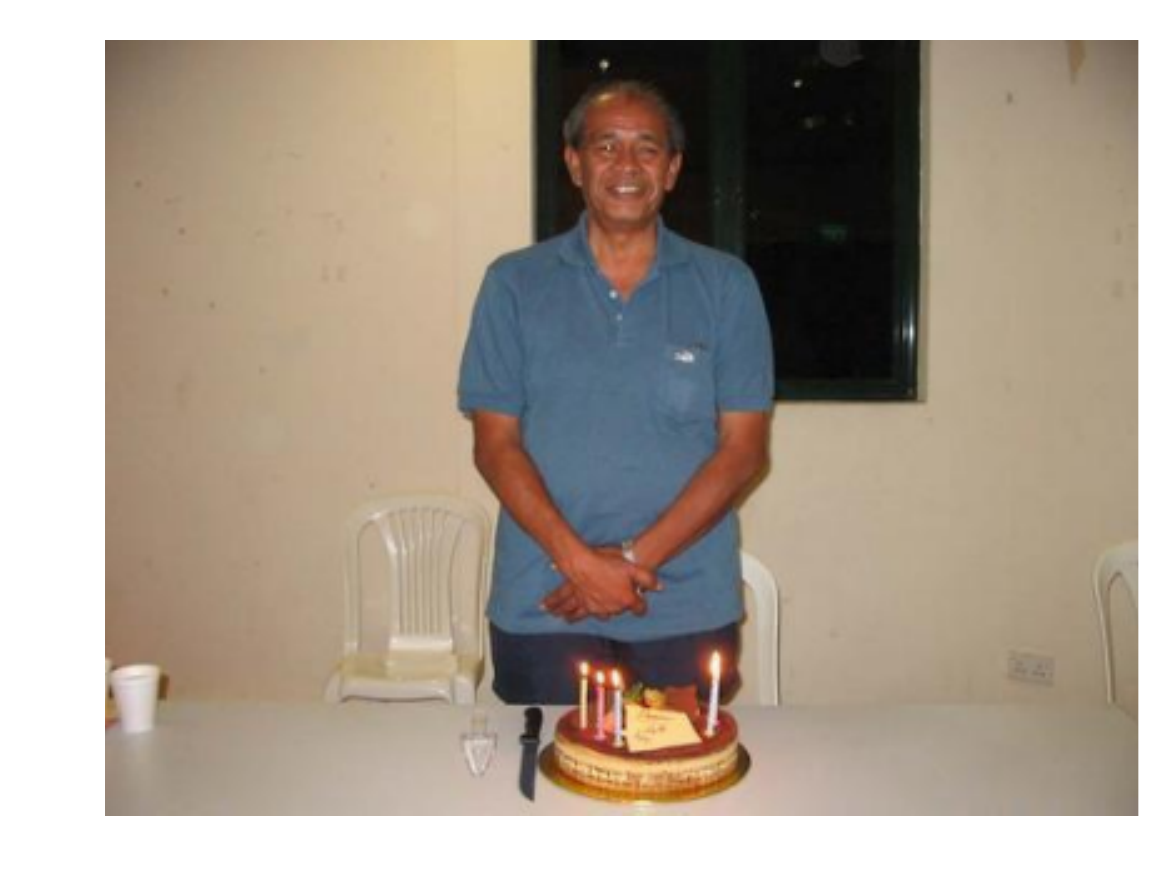}
\end{subfigure}
\begin{subfigure}[b]{0.19\linewidth}
\includegraphics[width=\linewidth]{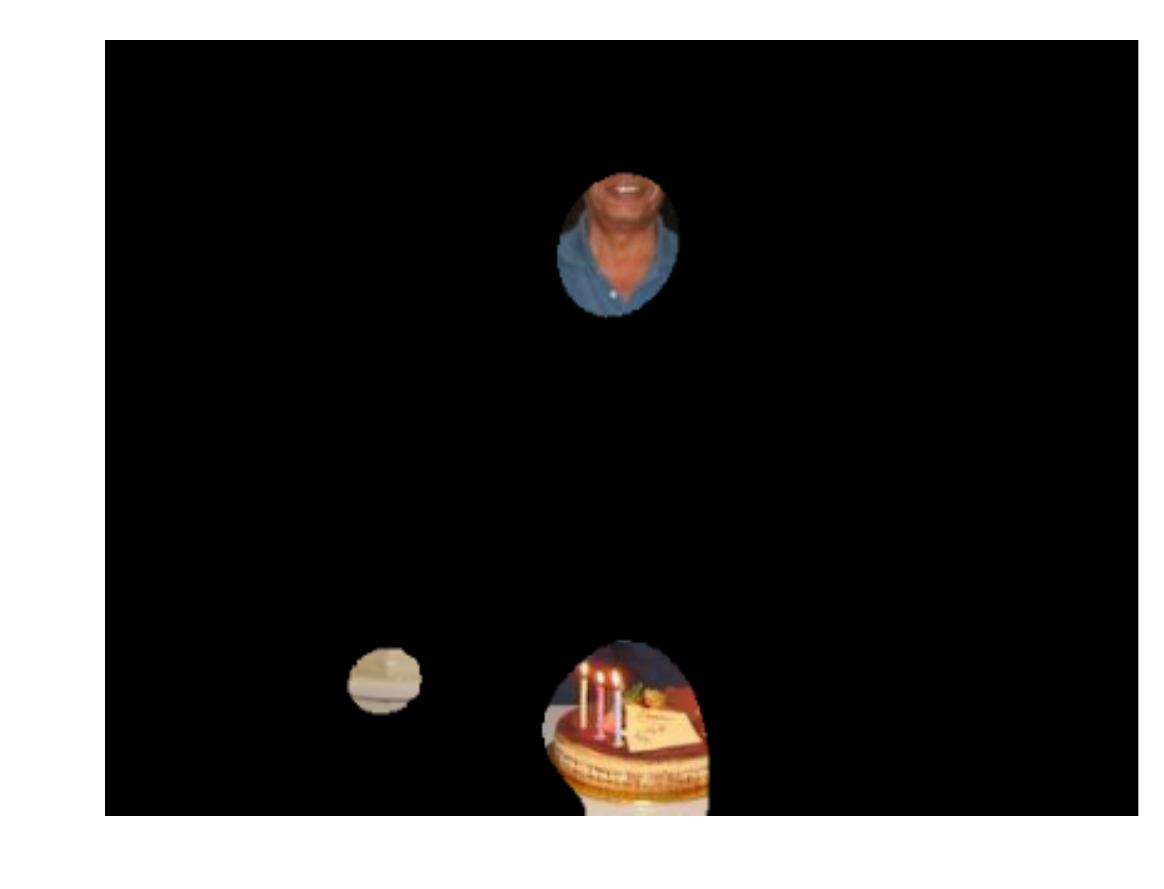}
\end{subfigure}
\begin{subfigure}[b]{0.19\linewidth}
\includegraphics[width=\linewidth]{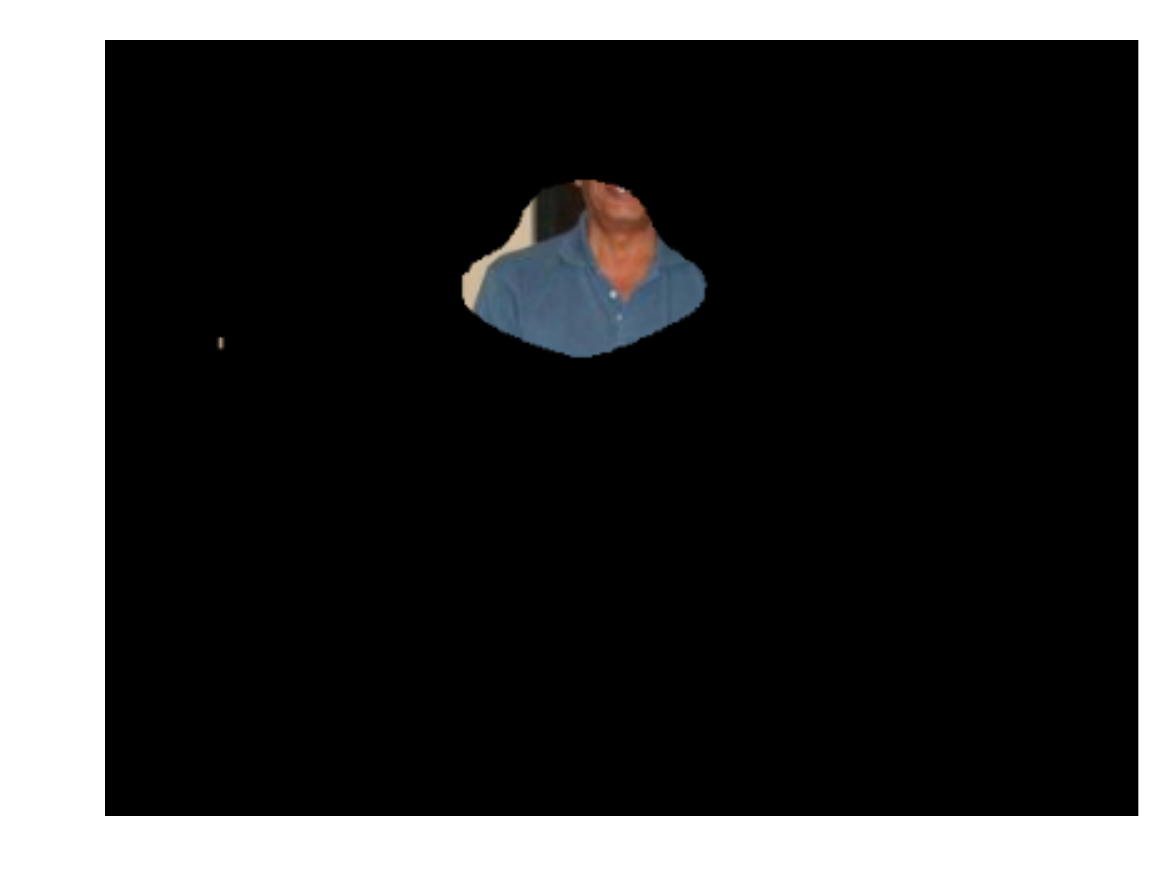}
\end{subfigure}
\begin{subfigure}[b]{0.19\linewidth}
\includegraphics[width=\linewidth]{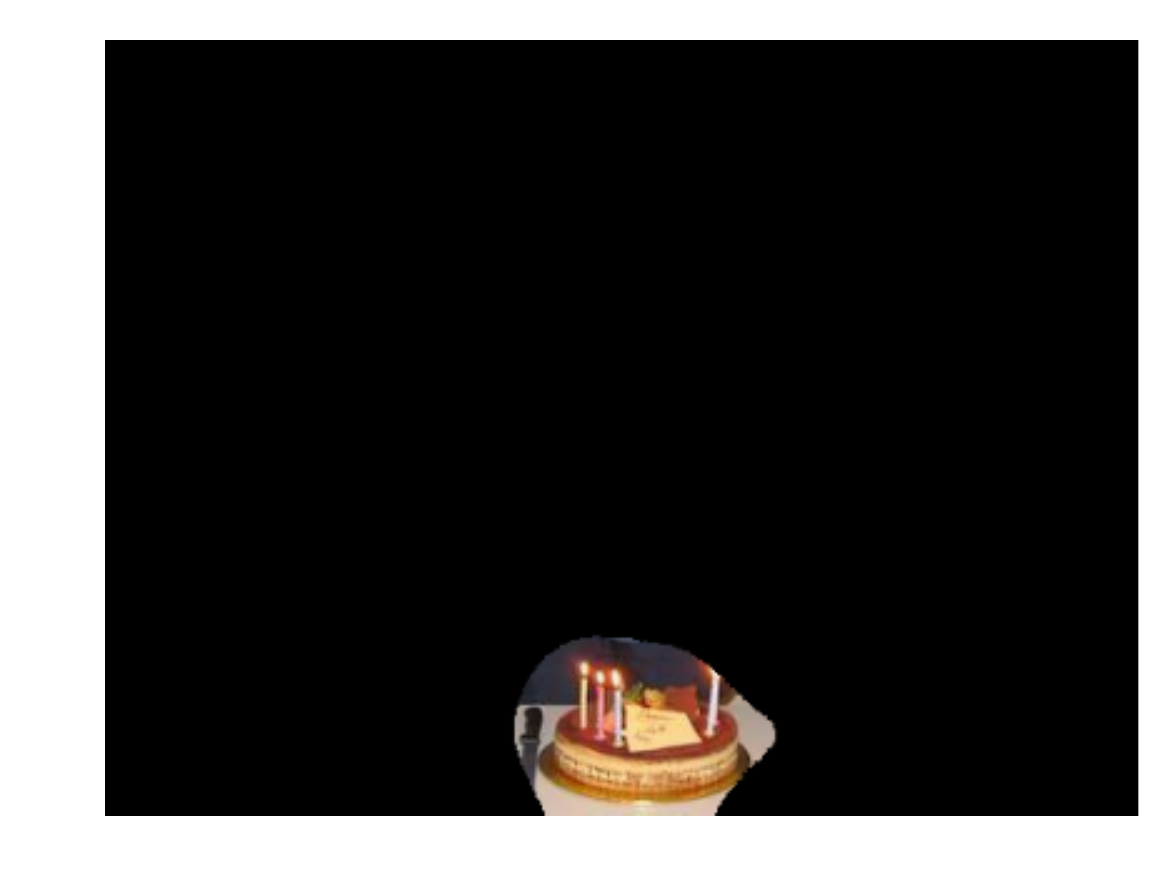}
\end{subfigure}
\begin{subfigure}[b]{0.19\linewidth}
\includegraphics[width=\linewidth]{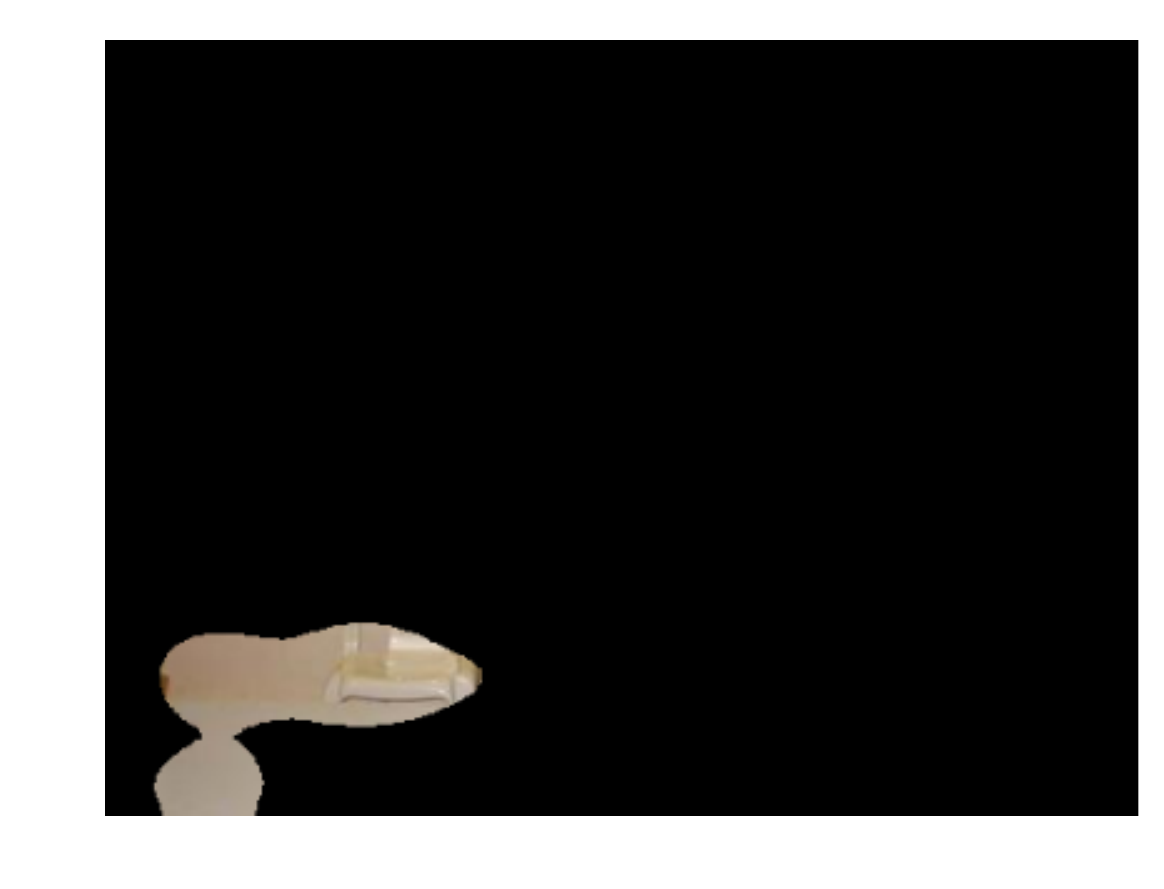}
\end{subfigure}
\caption{Example of human attention under captioning task and VGG-16's attention. From left to right: image, human attended regions, and VGG-16 attended regions that best correlated with each human attended region.}
\label{fig:hma}
\end{figure}

To answer these questions, we first transform the collected fixation data into saliency maps by convolving them with a Gaussian filter (sigma corresponding to one degree of visual angle in our experiments). Then, we threshold the saliency map by its top 5\% value and extract the connected regions. We then check how well the activation maps in the CNN, here layer conv5-3 of the VGG-16~\cite{simonyan2014very} (including 512 activation maps) correspond to the connected regions. To this end, for each connected region, we identify if there is an activation map that has a NSS score~\cite{bylinskii2018different} higher than a threshold (here T=4) within that connected region. If there exists one, then the corresponding connected region is also attended by the CNN. We report how many regions in images attended by humans are also attended by machine, as well as the mean highest NSS score of all the connected regions in all images (each connected region has a highest NSS score from 512 activation maps). We use fixation maps from free-viewing attention (free), first 3s fixations under the captioning task (cap3s), and the fixations of the whole duration of the image captioning task (cap). Results are shown in Table \ref{tab:hma_sta}. It can be seen that there exists a large agreement between internal activation maps of the encoder CNN and the human attended regions (over 70\%). Interestingly, despite not fine-tuning the CNN for captioning, this agreement is higher for the task-based eye movement data than that for free-viewing fixations (See example in Fig.~\ref{fig:hma}).

\begin{table}[t!]
\small
\caption{Attention agreement between human and the visual encoder (pre-trained CNN).}
\centering
\begin{tabular}{c|cc}
\hline
 & percentage & mean value\\
 \hline
 \hline
 free &72.5\% & 5.43\\
 cap3s &78.1\%& 5.62\\
 cap &77.9\% &5.61\\
 \hline
 \end{tabular}
 \label{tab:hma_sta}
 
\end{table}

\subsubsection{Attention in image captioning models}
How well does the top-down attention mechanism in the automatic image captioning model agree with human attention when describing images? We study the spatial and temporal consistency of the soft-attention mechanism in~\cite{xu2015show} with human attention in image captioning.

\begin{table}[t]
\small
\caption{Spatial attention consistency evaluation for bottom-up saliency model (SalGAN), and top-down attention captioning model (Soft-attention). Evaluated by NSS/s-AUC.}
\vspace{-15pt}
\centering
\begin{tabular}{cc|c |c}
   &  &\multicolumn{2}{c }{Ground truth}\\
\cline{3-4}
& Model &free-viewing &image captioning\\
\hline
\hline
&SalGAN &1.929/0.72&1.618/0.677\\
 & Soft-attention &1.149/0.622 &1.128/0.622\\
\hline
\end{tabular}
\label{tab:spa_att}
\end{table}

\paragraph{Spatial consistency:}
We assess the consistency between the spatial dimension of human attention and machine. 
For machine, the spatial attention is computed as the mean saliency map over all the generated words. 
We compute the NSS and s-AUC~\cite{bylinskii2018different} over this saliency map using human fixations.
We also compare with bottom-up saliency models by computing the NSS and s-AUC over the saliency maps of SalGAN~\cite{pan2017salgan}, a leading saliency model without centre-bias.

\begin{figure}[t]
\centering
\begin{subfigure}[b]{0.19\linewidth}
\includegraphics[width=\linewidth]{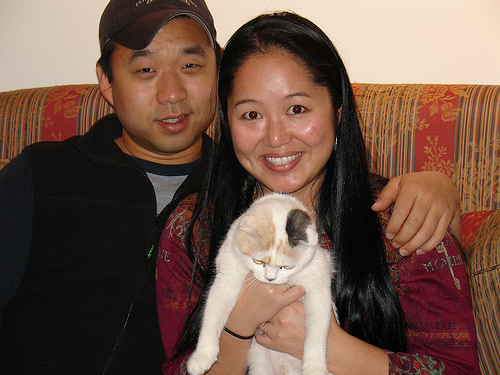}
\end{subfigure}
\begin{subfigure}[b]{0.19\linewidth}
\includegraphics[width=\linewidth]{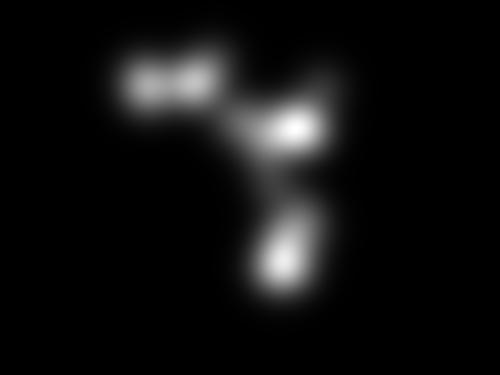}
\end{subfigure}
\begin{subfigure}[b]{0.19\linewidth}
\includegraphics[width=\linewidth]{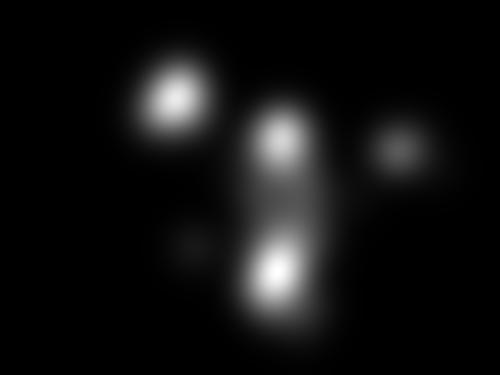}
\end{subfigure}
\begin{subfigure}[b]{0.19\linewidth}
\includegraphics[width=\linewidth]{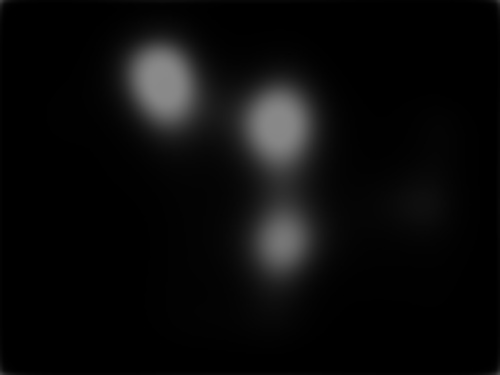}
\end{subfigure}
\begin{subfigure}[b]{0.19\linewidth}
\includegraphics[width=\linewidth]{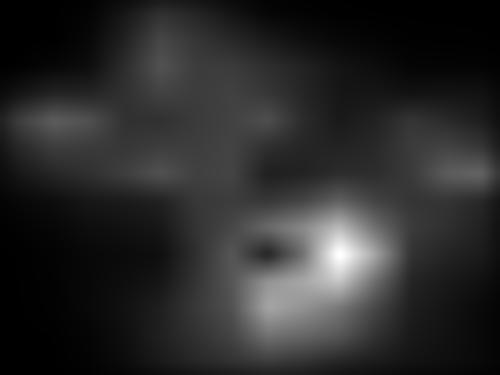}
\end{subfigure}
\begin{subfigure}[b]{0.19\linewidth}
\includegraphics[width=\linewidth]{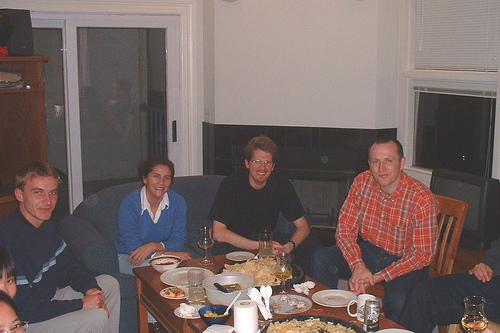}
\end{subfigure}
\begin{subfigure}[b]{0.19\linewidth}
\includegraphics[width=\linewidth]{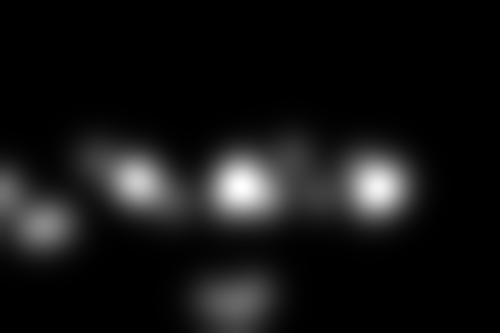}
\end{subfigure}
\begin{subfigure}[b]{0.19\linewidth}
\includegraphics[width=\linewidth]{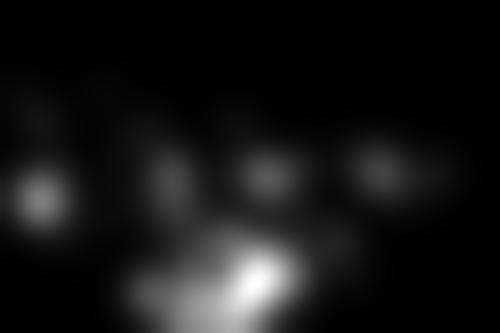}
\end{subfigure}
\begin{subfigure}[b]{0.19\linewidth}
\includegraphics[width=\linewidth]{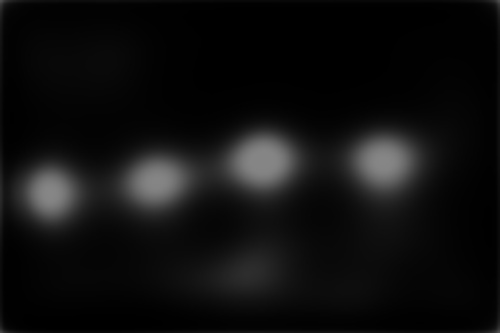}
\end{subfigure}
\begin{subfigure}[b]{0.19\linewidth}
\includegraphics[width=\linewidth]{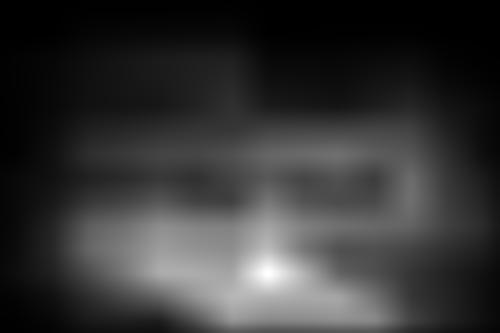}
\end{subfigure}
\caption{Example of spatial attention difference. From left to right: original image, attention in free-viewing, attention in image captioning, saliency map predicted by SalGAN, and saliency map from top-down image captioning model.}
\label{fig:sal_td}
\end{figure}

Table~\ref{tab:spa_att} summarizes the consistency of the saliency maps generated by a standard bottom-up saliency model (trained on free-viewing data) \cite{pan2017salgan} and a top-down soft attention image captioning system \cite{xu2015show}, with ground truth saliency maps captured either in the free-viewing or the captioning condition (full duration). Interestingly, the bottom-up saliency obtains higher scores on both free-viewing and task-based ground-truth data. In other words, a bottom-up model 
is a better predictor of human attention than the top-down soft-attention model, \textit{even for the captioning task.} Fig.~\ref{fig:sal_td} illustrates some example maps.

\begin{figure}[h]
\centering
\begin{subfigure}[t]{0.1\linewidth}
\includegraphics[width=\linewidth]{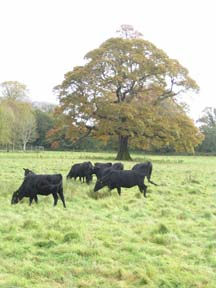}
\end{subfigure}
\begin{subfigure}[b]{0.88\linewidth}
\includegraphics[width=\linewidth]{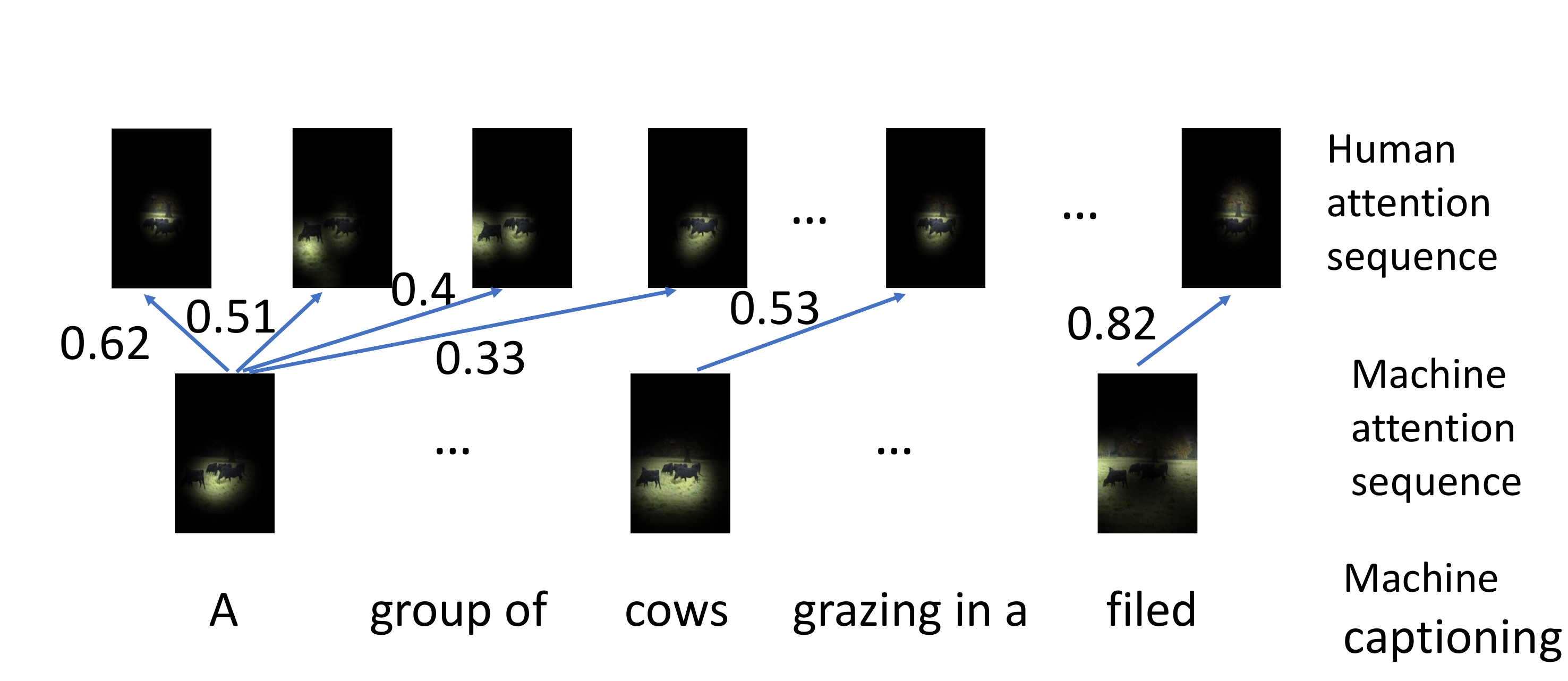}
\end{subfigure}
\caption{Example of Dynamic Time Warping between human attention and machine attention. Top row is the human attention sequence on the image when describing the image, the bottom row is the top-down model's attention sequence when generating the caption for the image, the number besides each blue arrow is the distance for each warping step.}
\label{fig:dtw}
\end{figure}

\begin{figure}[h]
\centering
\begin{subfigure}[b]{0.48\linewidth}
\includegraphics[width=\linewidth]{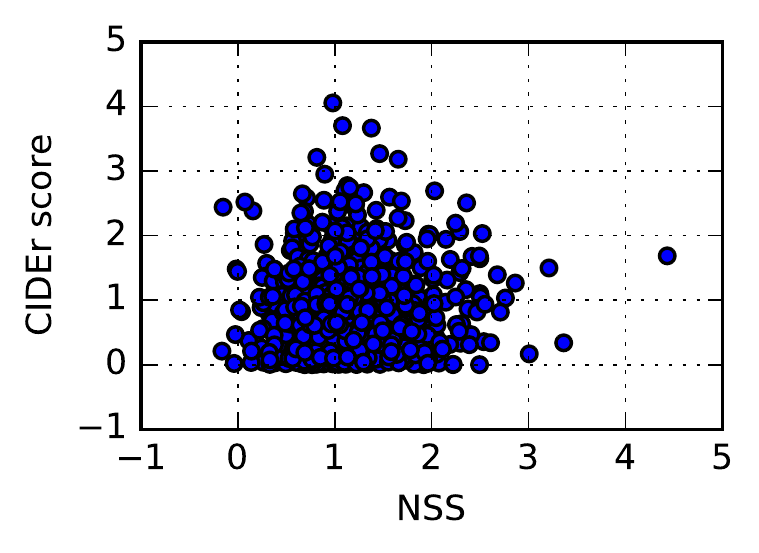}
\end{subfigure}
\begin{subfigure}[b]{0.48\linewidth}
\includegraphics[width=\linewidth]{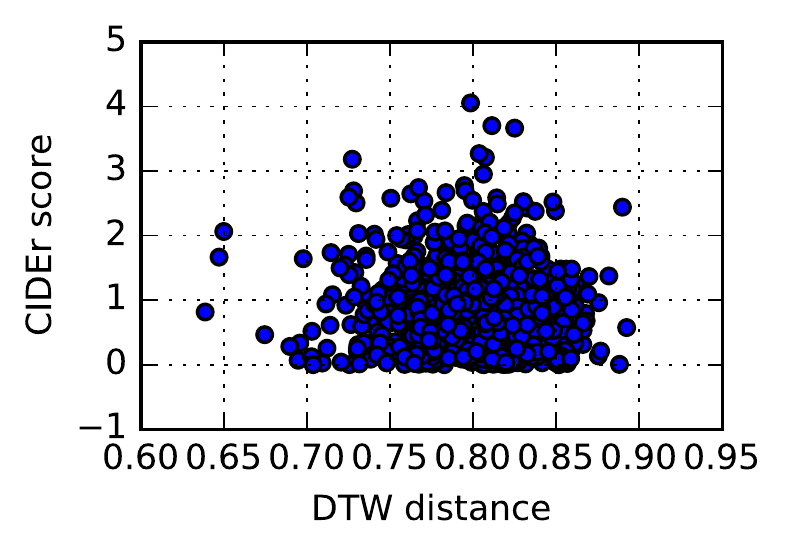}
\end{subfigure}
\vspace{-7pt}
\caption{Correlation between machine-human attention congruency (spatial and temporal) and machine performance on image captioning (CIDEr score).}
\label{fig:cor}
\end{figure}

\vspace{-10pt}
\paragraph{Temporal consistency:} What is the temporal difference between human and machine attention in image captioning?
Here, for the human fixation data, we split the sequence of fixations by intervals of $0.5s$ using the recorded sample time stamps. The fixations of each interval are then transformed into separate saliency maps, resulting in a sequence of saliency maps. For machine attention, we use the sequence of generated saliency maps during the scene description. We, then, employ Dynamic Time Warping (DTW)~\cite{muller2007dynamic} to align the sequences and compute the difference between them. Fig.~\ref{fig:dtw} shows this process for an example sequence. We report the distance between each frame pair as
$1-\mbox{SIM}(h_i,m_j)$,
where $h_i$ is the $i^{th}$ frame in the human attention sequence, $m_j$ is the $j^{th}$ frame in the machine attention sequence, and SIM is the similarity score~\cite{bylinskii2018different} between the two attention maps. The final distance between two sequences is the total distance divided by the path length in DTW. Our analysis shows a mean difference of $0.8$, which is significantly large and demonstrates that the two attention patterns differ significantly over time. 

\vspace{-10pt}
\paragraph{Correlation between machine captioning performance and machine-human attention congruency:}
Is the consistency between the machine and human subjects' attention patterns a predictor of the quality of the descriptions generated by the machine? To answer this question, we compute the \textit{Spearman correlation coefficient} between the machine performance on each image instance in terms of caption quality (CIDEr score) and the consistency of machine attention (spatial and temporal) with human (NSS score for spatial consistency, DTW distance for temporal consistency). Results are visualized in Fig.~\ref{fig:cor}, indicating a very low coefficient, 0.01 and -0.05 for spatial and temporal attention, respectively. 
In other words, there seems to be no relation between the similarity of the machine's attention to humans' and the quality of the generated descriptions. 


\subsection{Can saliency help captioning?}
Based on the analytical result in Table~\ref{tab:pfe}, 96\% of described objects are fixated (87\% in free-viewing), which means the image saliency map provides a prior knowledge of where to attend in image captioning. In contrast, the soft-attention models for image captioning first treat all regions \textit{equally}, before \textit{re-weighting} each region when generating each word. 
\begin{figure*}
\centering
\includegraphics[width=0.85\textwidth]{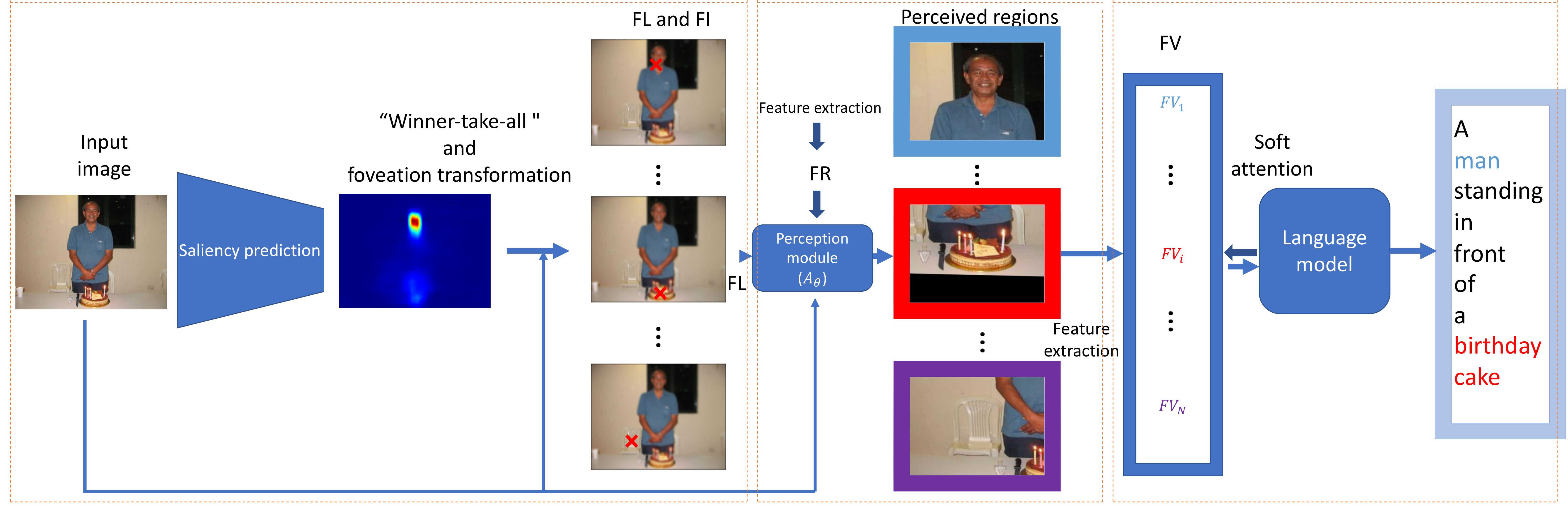}
\vspace{-15pt}
\caption{Architecture of the proposed method}
\vspace{-5pt}
\label{fig:archi}
\end{figure*}
Here, we check if image saliency can help image captioning by proposing a generic architecture, which combines the visual saliency and soft-attention mechanism for image captioning  as depicted in Fig.~\ref{fig:archi}. Our architecture has three parts: a \textit{saliency prediction module} (SPM), a \textit{perception module} (PM), and a \textit{language model} (LM). In the SPM part, we train a saliency prediction model on the \textit{capgaze2} corpus to predict a saliency map for each image\footnote{We adopt the model in~\cite{he2019understanding} for saliency prediction}. From this saliency map, we use a ``winner-take-all'' approach~\cite{koch1987shifts} to extract a set of fixated locations (FL) for each image. We denote those locations as:
\begin{equation}
FL=\{(x_1,y_1),\dotsi,(x_i,y_i),\dotsi,(x_N,y_N)\}
\end{equation}
For each fixated location in the image, we apply a foveation transformation~\cite{wang2017scanpath}, producing a set of foveated images (FI) for those fixated locations:
\begin{equation}
FI = \{FI_1,\dotsi,FI_i,\dotsi,FI_N\}
\end{equation}
We further process each foveated image with a pre-trained CNN, yielding a $K$ dimensional vector for each foveated image. Finally, for each image, we have a set of foveated representations (FR):
\begin{equation}
FR = \{FR_1,\dotsi,FR_i,\dotsi,FR_N\}
\end{equation}

The bridge between our SPM and LM is a learned perception module (PM), parameterized by a function $f$, in which we used a Localised Spatial Transformer Network~\cite{jaderberg2015spatial} (LSTN). For each fixated image location $(x_i,y_i)$, the PM generates an affine transformation ($A_i$), based on the corresponding $FR_i$, to perceive a region centred at the fixated location:
\begin{equation}
    \begin{split}
    A_i & =\begin{bmatrix}
    f(FR_i)&\aug&(x_i,y_i)^\top
    \end{bmatrix}\\
    &=\begin{bmatrix}
    \theta_{i_{11}}&\theta_{i_{12}}&\aug&x_i\\
    \theta_{i_{21}}&\theta_{i_{22}}&\aug&y_i\\
    \end{bmatrix}
    \end{split}
    \raisetag{2\normalbaselineskip}
\end{equation}
Each perceived region is then processed by a feature extraction network, and represented by a vector of dimension $K$. Finally, for each image, it has a set of feature vectors (FV):
\begin{equation}
FV = \{FV_1,\dotsi,FV_i,\dotsi,FV_N\}
\end{equation}
The LM is a LSTM with a soft attention module (parameterized by a learned function $att$). The soft attention module receives the FV as input. Based on the hidden state of LSTM ($\mathbf{h}$) and each feature vector in FV, LM generates a weight ($w$) for each feature vector and then takes the weighted sum of those feature vectors (WSFV) in FV to update the LSTM state and to generate the next word:
\begin{equation}
w_i = att(FV_i,\mathbf{h})
\end{equation}
\begin{equation}
WSFV =\frac{1}{N}\sum_{i=1}^{N} w_i\cdot FV_i
\end{equation}
The only difference between our model and the original soft attention model in~\cite{xu2015show} is that our attention module only emphasizes salient regions \textit{guided by the SPM and perceived by PM}, whereas the original soft attention model emphasizes \textit{all} regions in the image when generating each word.

Our architecture is trained in two stages. In the first stage, we train the SPM, and extract the FL and FR. Then, the PM and LM are trained jointly by minimizing the cross entropy loss of the caption generation. The pre-trained feature extraction for FV and FR is resnet-18~\cite{he2016deep}, which transform each foveated image and each perceived region into a 512 dimensional feature vector. The learning rate is set to $10^{-3}$, and is decreased by a factor of $0.8$ every 3 epochs. Early stopping is used if the BLEU-4~\cite{papineni2002bleu} score does not increase in five consecutive epochs. Our model is trained and tested on Flickr30k and MSCOCO benchmarks using the Karpathy \etal's split~\cite{karpathy2015deep}. 

Four metrics are used for evaluation, including BLEU-4 (B4), ROUGEL (RG)~\cite{lin2004rouge}, METEOR (MT), and CIDEr (CD). We also consider the use of free-viewing saliency in the image captioning (\ie saliency prediction model trained on Salicon~\cite{jiang2015salicon} database).

The performance of our architecture is shown in Table~\ref{tab:perfm1} and~\ref{tab:perfm2}. Our baseline model is the soft attention model in~\cite{xu2015show} (for fair comparison, we re-implement this model with resnet-18 as backbone~\footnote{Implemented using the code from: \url{https://github.com/sgrvinod/a-PyTorch-Tutorial-to-Image-Captioning}}). Our model significantly improves the performance of the soft-attention model by integrating a bottom-up saliency approach to the soft attention model. The model using \textit{task saliency} (saliency prediction model trained on our \textit{capgaze2} corpus) performs better than the one trained using free-viewing saliency---although the difference is not large. Our model is a general architecture, which could easily be integrated with other CNN backbones or the \textit{adaptive attention} mechanism in~\cite{lu2017knowing}. We also believe that the architecture can be applied to other tasks where visual saliency is important.
\begin{table}[ht]
\small
\caption{Performance on Flickr30k testing dataset (ours-free means saliency prediction model trained on free-viewing saliency database)}
\vspace{-5pt}
\centering
\small
\begin{tabular}{c | c c c c}
\hline
    Model  &   B4   &  MT  & RG &CD\\
\hline
\hline
baseline(soft attention)     & 0.191       &  0.171    &   0.419 & 0.352 \\
ours-free    &  0.213        & 0.175      & 0.431 &0.403 \\
ours   &   0.22     & 0.184         & 0.441 &0.416\\
\hline
improvement &15.2\% &7.6\% &5.3\% &18.2\%\\
\hline
\end{tabular}
\label{tab:perfm1}
\end{table}

\begin{table}[ht]
\small
\caption{Performance on MSCOCO testing dataset}
\vspace{-5pt}
\centering
\small
\begin{tabular}{c | c c c c}
\hline
    Model  &   B4   &  MT  & RG &CD\\
\hline
\hline
baseline(soft attention)     & 0.281       &  0.223    &   0.496 & 0.81 \\
ours-free    &  0.297        & 0.234      & 0.511 &0.889 \\
ours   &   0.303     & 0.238         & 0.518 &0.907\\
\hline
improvement &7.8\% &6.7\% &4.4\% &12\%\\
\hline
\end{tabular}
\label{tab:perfm2}
\vspace{-10pt}
\end{table}

\section{Discussions and Conclusion}

In this paper, we introduced a novel, relatively large dataset consisting of synchronized multi-modal attention and caption annotations. We revisited the consistency between human attention and captioning models on this data, and showed that human eye-movements differ between image captioning and free-viewing conditions. We also reconfirmed the strong relationship between described objects and attended ones, similar to the findings that have been observed in free-viewing experiments.

Interestingly, we demonstrated that the top-down soft-attention mechanism used by automatic captioning systems captures neither spatial locations nor the temporal properties of human attention during captioning. Also, the similarity between human and machine attention has no bearing on the quality of the machine generated captions. Finally, we show that attune soft attention captioning models to image saliency, demonstrating significant performance improvement to the purely top-down soft attention approach.

Overall, the proposed dataset and analysis offer new perspectives for the study of top-down attention mechanisms in captioning pipelines, providing critical hitherto missing information that we believe will assist further advancements in developing and evaluating image captioning models.

\noindent \textbf{Acknowledgements:} The authors thanks the volunteers for their help in the data collection. 
This research is supported by the EPSRC project
DEVA (EP/N035399/1). Dr Pugeault is supported by the Alan Turing Institute (EP/N510129/1).

{\small
\bibliographystyle{ieee_fullname}
\bibliography{egbib}
}

\end{document}